# Conscious AI

**Hadi Esmaeilzadeh[1]**
University of California San Diego
hadi@eng.ucsd.edu

**Reza Vaezi[1]**
Kennesaw State University
reza.vaezi@kennesaw.edu

## Abstract

Recent advances in artificial intelligence (AI) have achieved human-scale speed and accuracy for classification tasks. In turn, these capabilities have made AI a viable replacement for many human activities that at their core involve classification, such as basic mechanical and analytical tasks in low-level service jobs. Current systems do not need to be conscious to recognize patterns and classify them. [1]However, for AI to progress to more complicated tasks requiring intuition and empathy, it must develop capabilities such as metathinking, creativity, and empathy akin to human self-awareness or consciousness. We contend that such a paradigm shift is possible only through a fundamental shift in the state of artificial intelligence toward consciousness, a shift similar to what took place for humans through the process of natural selection and evolution. As such, this paper aims to theoretically explore the requirements for the emergence of consciousness in AI. It also provides a principled understanding of how conscious AI can be detected and how it might be manifested in contrast to the dominant paradigm that seeks to ultimately create machines that are linguistically indistinguishable from humans.

**Keywords**: artificial intelligence, AI, consciousness, service, AI consciousness

---

[1] Both authors have equal contributions and the order was decided by a coin toss.

# Introduction

Artificial Intelligence (AI) is rapidly becoming the dominant technology in different industries, including manufacturing, health, finance, and service. It has demonstrated outstanding performance in well-defined domains such as image and voice recognition and the services based on these domains. It also shows promising performance in many other areas, including but not limited to behavioral predictions. However, all these AI capabilities are rather primitive compared to those of nature-made intelligent systems such as bonobos, felines, and humans because AI capabilities, in essence, are derived from classification or regression methods.

Most AI agents are essentially superefficient classification or regression algorithms, optimized (trained) for specific tasks; they learn to classify discreet labels or regress continuous outcomes and use that training to achieve their assigned goals of prediction or classification (Toreini et al., 2020). An AI system is considered fully functional and useful as long as it performs its specific tasks as intended. Thus, from a utilitarian perspective, AI can meet its goals without any need to match the capabilities of nature-made systems or to exhibit the complex forms of intelligence found in those systems. As Chui, Manyika, and Miremadi (2015) stated, an increasing number of tasks performed in well-paying fields such as finance and medicine can be successfully carried out using current AI technology.  However, this does not stop AI from developing more complex forms of intelligence capable of partaking in more human-centered service tasks.

From the invention of the wheel to the industrial revolution and the current age of intelligent systems, different machines were invented and prevailed to assist us in achieving our desired outcomes more effectively and efficiently at different levels of complexity. In other words, machines prevailed because they served us reliably and consistently; they were in service to us. However, they have advanced so far in their capabilities and intelligence that we have begun to wonder whether one day they might be able to replace humans in complicated tasks that throughout history have been deemed exclusively in the human domain (Huang & Rust, 2018). They have come so far that people prefer AI-generated (machine-generated) recommendations over human recommendations for utilitarian decision making (Longoni & Cian, 2020). As AI continues to integrate with more human activities and exhibit exceptional utilitarian value in serving us, it is only natural to use natural intelligence as a comparative benchmark to understanding existing AI technologies and to use these comparisons to predict future AI progress (Stone et al., 2016).

Consequently, Huang and Rust (2018) identified four levels of AI intelligence and their corresponding service tasks: mechanical, analytical, intuitive, and empathic. Mechanical intelligence corresponds to mostly algorithmic tasks that are often repetitive and require consistency and accuracy, such as order-taking machines in restaurants or robots used in manufacturing assembly processes (Colby, Mithas, & Parasuraman, 2016). These are essentially advanced forms of mechanical machines of the past. Analytical intelligence corresponds to less routine tasks largely classification in nature (e.g., credit application determinations, market segmentation, revenue predictions, etc.); AI is rapidly establishing its effectiveness at this level of analytical intelligence/tasks as more training data becomes available (Wedel & Kannan, 2016). However, few AI applications exist at the next two levels, intuitive and empathic intelligence (Huang & Rust, 2018). Empathy, intuition, and creativity are believed to be directly related to human consciousness (McGilchrist, 2019). According to Huang and Rust (2018), the progression of AI capabilities into these higher intelligence/task levels can fundamentally disrupt the service industry, and severely affect employment and business models as AI agents replace more humans in their tasks. The achievement of higher intelligence can also alter the existing human-machine balance (Longoni & Cian, 2020) toward people trusting "word-of-machine" over word-of-mouth not only in achieving a utilitarian outcome (e.g., buying a product) but also toward trusting "word-of-machine" when it comes to achieving a hedonic goal (e.g., satisfaction ratings, emotional advice).

Whether AI agents can achieve such levels of intelligence is heatedly debated. One side attributes achieving intuitive and empathic levels of intelligence to having a subjective biological awareness and the conscious state known to humans (e.g., Azarian, 2016; Winkler, 2017). The other side argues that everything that happens in the human brain, be it emotion or cognition, is of computational nature at the neurological level. Thus, it would be possible for AI to achieve intuitive and empathic intelligence in the future through advanced computation (e.g., McCarthy, Minsky, Rochester, & Shannon, 2006; Minsky, 2007). We contend that conscious AI, capable of empathic intelligence, might be possible without the need for biological processes. In this paper, we focus on what is required for such a conscious state to arise and how we can recognize conscious machines. We stop short of proposing the detailed, underlying computational models that may lead to a conscious AI.

Consequently, this paper puts forward a theory on how consciousness would emerge in its primitive state in AI agents and how it may progress toward a point that we can deterministically recognize as conscious AI. We hope this theory will chart a route toward exploring a possible path toward empathic machines. For this purpose, in addition to the AI in the service literature, we rely on and contribute to two other research traditions; one aims to detect intelligence in machines, and the other aims to understand consciousness. This paper contributes to the service literature by advancing our understanding of empathic AI as the final stage of intelligence, as defined by Huang and Rust (2018). It also provides principled guidance towards the detection of machines' leap to such conscious states. Understanding such a transition will alter our conceptions of service because it will change the human-machine balance, affecting the service job markets.

According to Moor (2003), the research on detecting intelligence in machines starts with Alan Turing's seminal work (Turing, 1950) and is defined by it through constant efforts to pass the proposed test. Turing (1950) introduced a test that later became known as the Turing Test to pinpoint when it can be said a machine (a standard computational machine) is capable of thinking. As we know it, the test will determine if a machine is *linguistically indistinguishable* from humans (Bringsjord & Govindarajulu, 2018). Despite the most recent advances in the field of natural language processing (Heaven, 2020) and *claims* reported by (Humphrys, 2009) that AI has already achieved language indistinguishability, we are not aware of any such claims that are independently verifiable and free of controversies and special conditions. Furthermore, the Turing Test can identify "thinking machines" only at their maturity when they have achieved linguistic indistinguishability from humans. In contrast, our proposed theoretical perspective contributes to AI research by providing a principled framework that can identify "thinking machines" through their path toward maturity along with minimum requirements under which machine consciousness can emerge.

Research on consciousness is vast and involves many scientific disciplines (e.g., psychology, sociology, neurology, pathology, philosophy, physics, etc.). Defining consciousness from the perspectives of various disciplines requires diverse expertise and can result in volumes of books. As a result, and for the purpose of this paper, we consider consciousness from philosophical and psychological (philo-physiological) and social-psychological perspectives. Additionally, although we all seem to know we are conscious, we cannot agree on the exact conception of consciousness (Goldstein, 2012). Thus, it is important to note we are not trying to address the hard problem of consciousness (Chalmers, 1996; Goldstein, 2012; McDermott, 2007) in defining what consciousness is and how it functions in humans or other beings (e.g., AI agents). We raise these issues neither to defend our perspective on consciousness nor to prevent criticism; we raise them to clarify this paper's scope and encourage future research to build on different ontological and epistemological perspectives.

We propose that consciousness in AI is an emergent phenomenon that manifests when two machines co-create their own language through which they can communicate their internal state of time-varying symbol manipulation, especially when these co-created symbols do not correspond to external objects and represent shared meta concepts. This prospective language will be a novel creation and is independent of human or any other conscious agent's influence. As such, consciousness in AI does not focus on classifying, detecting, recognizing, or conversing with humans as the current paradigm of AI and the Turing Test aim to achieve. In contrast, it is a process through which one AI agent becomes conscious of itself through another AI agent that cooperates in the creation of a common medium of communication to express internal states.

This insight contributes to consciousness research by bridging the gap between philo-psychological theories of consciousness and the interactionist social-psychological view of consciousness and by depicting *a* possible path toward machine consciousness. We contribute to the understanding of consciousness by indicating the existential requirements for its emergence. We also portray a general mechanism for the emergence of consciousness in AI agents and foresee the development of conscious AI.

The rest of this paper briefly reviews the literature on intelligent machines and considers prominent philosophical and psychological theories of consciousness. Then, it moves to explore consciousness from a social-psychological viewpoint. Next, it presents our theory of AI consciousness, discusses it in detail, and envisions a path toward conscious AI. The paper ends with the implications of this theory on the future of research and practice.

## Intelligent Machines

One might track the inception of intelligent machines to the invention of the Electronic Numerical Integrator and Computer—ENIAC— the first computing machine in 1943. Later, Alan Turing (Turing, 1950) suggested the possibility of a thinking computing machine in which the machine could achieve linguistic indistinguishability from humans. However, it was not until the mid-1950s that the field of AI took root as the result of a conference at Dartmouth College sponsored by the Defense Advanced Research Project Agency (DARPA). Since then, most of the philosophical discussions around AI, as well as the main goal of the AI field, have been influenced by Turing's vision of a thinking machine and his proposed test of language indistinguishability to recognize such machines (Bringsjord & Govindarajulu, 2018).

The ultimate aim of the AI research has been, and arguably still is, to create a machine capable of passing the Turing Test and then achieve functionalities comparable to and even indistinguishable from adult humans (Bringsjord & Govindarajulu, 2018; Brooks, Breazeal, Marjanović, Scassellati, & Williamson, 1998; Michie, 1973; Moor, 2003; Russell & Norvig, 2016; Yampolskiy, 2013). Advances in AI, including IBM Watson's victory (Ferrucci, 2012) in the open domain question-answering game of Jeopardy on television, prompted some people to argue that the Turing Test has been passed (Castelluccio, 2016). However, as impressive as this achievement is, Watson cannot converse on the fly about any given subject, a capability implied in the Turing Test (Bringsjord & Govindarajulu, 2018; Marcus, 2013). More recent developments in AI have come closer to passing the Turing Test through the OpenAI, GPT-3 project, which is capable of automatically generating meaningful texts in the style of a given author, but it still lacks the live conversation feature proposed by this test (Brown et al., 2020; Heaven, 2020). We believe such a single-pointed focus on achieving language indistinguishability for AI agents has kept us from exploring other possibilities for conscious AI.

Some researchers have seen the Turing Test's exclusive focus on achieving linguistic indistinguishability as myopic and called for moving beyond this test and consider philosophical and theoretical frameworks that encompass a complete range of human cognitive functions (Harnad, 1991; Russell & Norvig, 2016), such as cooking, playing sports, and understanding and expressing emotions. Such concerns have resulted in the conception and differentiation of Weak AI and Strong AI. Weak AI is defined as information-processing machines that appear to possess the full range of human cognitive abilities, while Strong AI is defined as intelligent machines that actually possess all mental and comparable physical capabilities of humans, including phenomenal consciousness through advanced computation (Bringsjord & Govindarajulu, 2018). The possibility of Strong AI has many fierce critics who consider it impossible. Anchoring their arguments in a philosophical and physiological understanding of individual consciousness, they contend machines can never develop phenomenal consciousness through advanced computation (e.g., French, 2012; Penrose, 1994; J. Searle, 2014; J. R. Searle, Dennett, & Chalmers, 1997). We argue that a form of Strong AI with an emergent consciousness is possible. Because we are proposing a possible path toward conscious machines and are uninterested in addressing the hard problem of defining what constitutes consciousness, we will not dive further into the literature and arguments for either side of this debate.

## Consciousness Theories

Unlike pathological and neurological theories of consciousness that try to explain consciousness from physical and biological viewpoints, philo-psychological theories attempt to understand and explain consciousness from more abstract perspectives related to the mind's function. This section summarizes prominent theories of consciousness at the intersection of philosophy, psychology, and sociology. We start with a series of theories focusing on the mind, its states, and how it is related to its environment. Then, before outlining our theory of conscious AI, we move into the social-self theory of consciousness (Mead, 1913), which views consciousness as a social phenomenon instead of an individual one.

## Philo-Psychological Theories of Consciousness

These philo-psychological theories are primarily concerned about what consciousness is and how it comes to be within a conscious entity (a human). The structure of the mind, mental states, and how information is processed, retrieved, and stored in the mind all play significant roles in these theories. They are mostly focused on the internal processes related to consciousness while acknowledging the entity's environment as a source of stimuli for internal processing but not giving it a central role.

The representationalism school (both first-order and higher-order) reduces consciousness to mental representations of external objects such as photos, signs, natural objects, and their qualities (e.g., trees, their barks, branches, leaves, rings inside their barks, etc.). Mental representations can be either intentional or phenomenal. Intentional mental states (representations) are the ones that contain representational consents about something or directed at some object, such as when a conscious entity has thought about a laptop computer or a perception of a booklet. This kind of mental activity forms an intentional representation of the object in the entity's mind. On the other hand, phenomenal states such as joy, pain, and experiences of sound and color are emergent as opposed to intentional states. It is important to note that most conscious experiences contain both states, such as the visual and auditory perception of a video clip (Gennaro, 2015).

First-order representationalism is concerned with "what it is like" for an entity to be conscious. It holds that experience is transparent; hence phenomenal states of an experience can be explained as intentional states of such experience. For example, when an entity looks at a tree, what it is like for it to have a conscious experience is identical to the experience's representation of the tree (Dretske, 1997; Tye, 1995, 2000). This view asserts that an entity is conscious if it is capable of constructing direct representations of objects and their associated phenomenal states (Tye, 1995, 2000). The theory explains what a conscious experience is and contends that this experience is not separate from the experience's representation within the entity's mind (Herman, 1990; D. M. Rosenthal, 1993). However, this theory fails to explain mental states that exist but are not conscious, such as a toddler's mental states while dealing with objects.

Higher-order representationalism tries to address the shortcomings of first-order representationalism by differentiating between conscious and unconscious mental states. In short, this theory contends that a mental state (e.g., perception, thought) is conscious only if it becomes the object of a higher-order representation (Gennaro, 2011; Gennaro, 2015). In other words, a mental state is only considered conscious when another mental state within the same conscious entity is aware of it (Gennaro, 2011; Rosenthal, 1997). Here the state of awareness (e.g., attention) is a higher-order mental state that is directed toward the mental state that contains an experience or its representation. For example, an entity's desire to express its opinion becomes conscious only when the entity is aware of such a desire. Higher-order representationalism asserts that a conscious mental state emerges when one mental state (higher-order) is directed at another mental state of immediate representation or thought (Gennaro, 2011; Rosenthal, 2000). This theory defines consciousness as an emergent phenomenon that requires a kind of relationship between two mental states. Both first- and higher-order representationalism theories focus on the mechanism by which consciousness comes to be or exists within an entity's mind.

The self-representational theory of consciousness resembles higher-order representationalism in that they both propose that consciousness is generated as a result of the relationship between different orders of mental states (Kriegel, 2009; Rosenthal, 1986; Sartre, 1956). However, the self-representational theory suggests the higher-order state is a part of an overall and complex conscious mental state as opposed to this state being a distinct and independent state of itself (Gennaro, 2015). For example, when an entity desires a meal, this conscious mental state represents both the meal and itself. Hence, a mental state is conscious because it represents and reflects itself (e.g., one part of the state represents another part).

Multiple drafts theory rejects the notion of a specific location in an entity's brain or a state of mind (e.g., self) in which everything blends together to create consciousness (Dennett, 1991). Instead, it states that all kinds of mental activities occur concurrently in a conscious entity's mind. Over time the interpretations and frequent revisions of these activities will lead to the creation of a "center of narrative gravity" that forms the core of consciousness experience (Dennett, 2005). Accordingly, consciousness arises from multiple interpretations and revisions of an entity's experience with its environment.

Global workspace theory, similar to multiple drafts theory, is concerned with the process by which consciousness emerges. In providing a mental model of how the mind works (Baars, 1993, 1997), it depicts the mind as a global workspace (e.g., a blackboard, a theater) in which unconscious mental states and processes compete for the spotlight of an entity's attention and focus. The information under the spotlight then becomes available globally throughout the system (Baars, 1997; Shanahan, 2010; Taine, 1872; Wiggins, 2012). Accordingly, in this view, consciousness is created through global access to different information available in the entity's "nervous" system and offers a blended view of pure philosophical models of consciousness and those rooted in natural sciences (Shanahan, 2010).

A review of prominent philo-psychological theories of consciousness suggests these theories do not offer external observers a direct way of knowing whether they are dealing with a conscious entity. Instead, these theories are primarily concerned with what consciousness is and how it comes to be for an entity through an internal state and with the processes related to its origination. Possession of an internal state seems to be the implicit requirement these theories impose on conscious entities. However, as important and intuitive as having an internal state is for consciousness, it cannot guide us further, from the standpoint of an independent observer, in recognizing such a state in potentially conscious entities. Relying on these theories, a conscious entity itself may know it is conscious. However, an independent (external) observer may not be able to form an accurate conclusion about an entity's consciousness—unless the entity itself informs the observer of its conscious state. As a result, a theory is needed that can potentially offer external observers a way to examine and understand whether an entity (e.g., AI agent) can be considered conscious.

## The Social-Self Theory of Consciousness

The social-self theory of consciousness does not interpret consciousness as an individual phenomenon. Instead, consciousness is considered a social phenomenon; thus, individual conscious agents' mental states and processes are of little concern. Rather, the theory is focused on individual acts within a social context. Thus, an active social unit must be involved for consciousness to emerge; in other words, an environment must exist within which actors communicate and interact with each other. From this perspective, the sense of self and the main instrument of consciousness is a social phenomenon. It does not exist outside of a social matrix of social acts (Mead, 1913; Mead, 1934; Percy, 1958).

Although the idea of consciousness being a social phenomenon (Mead, 1913) predates philosopher Thomas Nagel's famous conceptualization of consciousness (Nagel, 1974), it essentially relies on a fundamental notion that an organism is conscious when it knows *what it is like to be* another organism. According to Nagel (1974, p. 440), *what it is like* does not mean "what (in our experience) it resembles," but rather "how it is for the subject himself." Consequently, the subject of inquiry is no longer the whatness (phenomenology) of consciousness but its "howness," as in how another entity perceives it. Thus, other conscious organisms must exist within the conscious entity's immediate environment to make it possible for the conscious organism to experience what it is like to be the other. These organisms form a social unit, and their interactions create a matrix of social acts.

In Mead's (1934) idea, consciousness requires actors capable of communicating with each other through an exchange of gestures (social acts) as the most primitive enablers of communication. Gestures primarily serve as a stimulus to other organisms involved in the same social act. For example, when a dog is ready to attack in a dog fight, its readiness serves as a gesture, a stimulus, to the other dog to respond accordingly with a gesture of its own. This exchange of gestures then leads to attitude adjustments in each dog, and this cycle continues until the situation is over.

Gestures, however, can be unconscious. We cannot deterministically say whether a dog is conscious of its gestures or of the adjustments of its gestures in a dog fight. Unconscious gestures are even evident in humans. For example, people who jump out of their seats and run away after hearing a loud sound may not be instantly aware of their reaction to such stimuli. When actors are fully aware of the meaning of their gestures, then we have a symbol. In other words, a symbol is a gesture that carries an explicitly shared meaning for all actors involved in a social act. Symbols arouse the same meaning in their initiator's mind as they do in their receiver's mind and create a shared understanding.

One can assume a person intends harm if that person approaches with a clenched fist. In this case, the clenched fist can be considered a symbol that carries the same meaning for involved actors. The one on the receiving side of the clenched fist assumes the imminence of an attack, while the initiator, who has had the same or a very similar experience, is implicitly ready to respond to the victim's defensive act. This back-and-forth exchange of symbols constitutes a matrix of social acts or a social matrix (Mead, 1913; Mead, 1934). A continuous exchange of symbols in an existing social matrix then constitutes a language. In humans, vocal gestures have mostly turned into symbols and created a variety of complex human languages (Baldwin, 1981; Mead, 1934).

Accordingly, consciousness in social-self theory requires a social matrix consisting of social acts and the exchange of symbols that lead to the creation of a language. Even though this theory is not concerned with mental processes, the existence of a mind, or an internal state, is implicitly assumed for actors. It also requires the initiator of a symbol to be able to respond to its own symbol implicitly because the other actor may respond explicitly. This requirement further reinforces the need for actors to have an internal state or a mind to respond to their own symbols internally in the same way that the other actors may respond externally.

Similar to the philo-psychological theories, the social-self theory faces the same issue of determining a given entity's conscious state. The entity might know it has an internal state, but external observers would be unable to infer such an internal state unless the entity itself stated that it had an internal state of such and such quality. The review of the literature revealed that neither the philo-psychological theories nor the social-self theory alone could attest to the existence of consciousness. None of them alone can provide guidance in positively determining whether an entity is conscious. Additionally, social-self theory does not concern itself with the creation of language. We, on the other hand, posit that the co-creation of the language is one of the missing links to consciousness.

## A Theory of AI Consciousness

This section uses AI, AI agent, and machine interchangeably, and in using them, we invariably refer to a standard hardware and software package that includes computational processing power, temporary and permanent memory structures, input and output ports, data storage, and computational capabilities. Moreover, to provide more clarity early on, we state our propositions before developing their supporting arguments.

**Proposition 1:** For consciousness to emerge, *two* AI agents capable of *communicating* with each other in a *shared environment* must exist.

To develop a clearer understanding of the consciousness phenomenon and devise the minimum requirements for the emergence of consciousness, we have adopted the assumed or given aspects of philo-psychological theories of consciousness and the social-self theory, namely the existence of "the other" and existence of "internal states", as explicit requirements for our theory. According to Mead (1934), language in the form of vocal symbols provides the mechanism for the emergence of consciousness. However, Mead does not concern himself with the creation of language. His view relies on the *existence of language* and its necessary elements (exchange of vocal symbols through social acts in a social matrix) as a required mechanism for consciousness. We, on the other hand, focus on the inception and development of language, in any possible form, by AI agents as a sign of emergent consciousness.

Language, in its essence, is a means of social interaction and a social phenomenon. It cannot be created in isolation when only one conscious entity exists within a given environment. Hence, this proposition suggests for an AI conscious state to emerge, we need at least two AI agents capable of communicating within a given environment to foster the creation of an AI-specific language. It should be noted that the properties of such communications (emergence and meaning) are fundamental to this theory, *because communicating computing machines already exist*, and yet they are not conscious. The following proposition is concerned with the emergent property of such communication.

**Proposition 2:** For consciousness to emerge, AI agents must exchange novel signals.

We envision consciousness as an emergent phenomenon and argue that the creation of a mechanism by which it emerges, namely language, should also be an emergent phenomenon representing the emergence of consciousness. By definition, an emergent phenomenon comes into being (emerges) without prior existence because of the interaction among parts of a whole (O'Connor, 2020). To infer emergence as a property of an existing system, one must observe "something new, a fresh creation" that emerges from the system instead of being the result of the system's working (Alexander, 1920; O'Connor, 2020). Thus, creation is embedded in the notion of emergence. To be precise, creation, in this context, refers to a spontaneous idea that appears without much deliberation instead of the creativity inherent in deliberate problem-solving activities. This kind of creation is essential in musical improvision, not composition (Hutchins & Hazlehurst, 1995; Plotkin, 1997; Wiggins, 2012). Accordingly, when detected in an AI agent's communication, the spontaneous (random) creation of signals can be considered an early signal of the emergence of consciousness. However, the emergence and detection of novel signals in communication between two machines is not enough by itself to indicate the existence of a conscious AI state. The detected novel signals should convey a shared meaning, which is the focus of the next proposition.

**Proposition 3:** For consciousness to emerge, AI agents must turn novel signals into symbols.

Today, most computers and AI agents are capable of communicating with each other, at least through some form of electronic signal or information packet exchange. Some novel communication signals, typically dismissed as errors, even exist between machines. However, this exchange of signals does not make machines conscious because no deterministic meaning is attached to these signals. Thus, the current exchange is considered unconscious and nothing more than the mere exchange of noise. For consciousness to emerge and to be observed by an independent onlooker requires more than a mere exchange of signals: there must be an exchange of meaning through signals. To preserve the consistency of terminology across various disciplines, we call these meaningful signals *symbols*. These symbols are going to be the building block of an AI-specific language.

We posit that the creation of symbols requires an agreement between two AI agents on the symbols' meaning. However, what symbol is associated with which object is, to a large degree, immaterial. What matters is the agreement between machines to use the same symbol for the same object or concept. For instance, the object tree is expressed as "arbor" in the Latin language and "дерево" in Russian. The symbol "arbor" has no material advantage or disadvantage over the symbol "дерево." What is important is that two agents have agreed to use "arbor" for the same object, a tree. Such agreement is the first step in turning a signal into a symbol by giving it a shared meaning. Thus, meaning arises from the agreement between agents, not from the symbol itself. For such an agreement to be reached, AI agents must have an internal state.

**Proposition 4:** For consciousness to emerge, AI agents must have an internal state.

For the shared meaning of a symbol to be inferred, Mead (1934) suggests conscious organisms should be aware of the meaning of their own signals and be able to respond to their own symbols implicitly through adjustment of their attitude as the other organism responds explicitly or implicitly. In other words, conscious organisms should take on the attitude of the other organisms toward their own initiated symbols implicitly as the other organisms respond explicitly. Similarly, we state that for an exchange of a meaningful signal or a symbol to be inferred, each AI agent should be aware of the meaning of its own symbols and be able to respond to its symbols implicitly as the other agent responds explicitly. This awareness and responsiveness implies that a conscious AI agent should have an internal state, known to itself to understand its own symbols and respond to them implicitly. The existence of an internal state as a prerequisite for the emergence of consciousness is also assumed in most of the philo-psychological theories of consciousness. In fact, the internal state and its processes are the sole focus of these theories (e.g., Dennett, 1991; Kriegel, 2009; Rosenthal, 1993; Tye, 2000).

Furthermore, an essential quality of a conscious entity implied in the social-self theory but explicitly stated by Nagel is to know what it is like to be the other (Nagel, 1974). For a conscious entity to know what it is like to be the other, it must have an internal state in which it can reconstruct the other. We believe conscious AI agents should not be an exception to this rule, and their internal state should enable them to know what it is like to be the other. Thus, we state that for consciousness to emerge in AI agents, they must possess an internal state.

**Proposition 5:** For consciousness to emerge, AI agents must communicate their internal state of time-varying symbol manipulation through a language that they have co-created.

For one machine's internal state to become known to another machine and eventually to an independent observer, we propose that a conscious AI agent (e.g., Machine A) must be able to communicate the contents of its internal state to the other AI agent (e.g., Machine B) through their mutually developed set of symbols. Such communication suggests the existence of an internal state, a mind, for AI agents and indicates the machines' ability to express and understand their internal states.

We envision three stages of development in the AI agents' path toward consciousness. In the first, the two machines should agree on a spontaneous (random) signal to represent a static (time-invariant) object in their environment. Once such an agreement is reached, the signal is turned into a symbol and must be moved into AI agents' permanent memory to be used in the future to refer to the same static object. The process creates a symbol comparable to what is commonly known as a noun in human languages. In the second stage, the two machines should agree on a random signal or set of previously created symbols to represent a dynamic (time-variant) concept related to an object in their environment. In other words, they should be able to describe the changing state of an external object.

An agreed-upon signal or a set of previously created symbols that refer to a decaying apple or a snoring cat in the machines' environment would be an example akin to the second stage of the AI agents' path toward consciousness. This is similar to the process of creating new verbs in human languages. In the final stage, the two machines should use a set of previously created symbols or a mixture of old symbols and novel signals to express their time-varying internal state of symbol manipulation. At this ultimate stage, in addition to their ability to refer to external objects and time-varying states of those objects, machines will also communicate their own internal states and how they manipulate symbols in real-time to create new symbols and their associated meanings. Once the third and final stage is observed, we can conclude consciousness has emerged in machines. We further explore this requirement in the next proposition.

**Proposition 6:** For the emergence of consciousness *to be concluded*, an onlooker should be able to observe two agents reaching an agreement about at least one of their state of time-varying symbol manipulation.

Proposition 5 may result in a language so different in terms of structure and form that is completely alien to us. For independent onlookers to conclude that we are observing conscious AI agents, we need to detect communications about their internal states and how those states change over time. To detect machines' communications about their internal states, we propose that independent onlookers should recognize an explicit agreement about the meaning of the communication. A good example would be two machines cooperatively completing a task they are not programmed to do. Completing a task in such a manner can point to active agreements in the communication of intent and time-varying internal states between machines.

## Discussion

## Service Implications

Conscious AI has many fundamental implications for our society. This concept can change how humans perceive AI and how they interact with AI-based technologies. It will necessitate new laws and bring forth concepts such as AI ethics and rights (Scheutz, 2016). Most importantly, it is going to affect the job market for humans, especially at the higher end of the job market in the service industry. From a utilitarian perspective, AI agents are viewed as service agents; they are built to help us with general (e.g., variety of personal assistants) or specific tasks (e.g., autonomous driving). In other words, they are of service to us (Colby et al., 2016; Huang & Rust, 2013, 2018). Conscious AI, by definition in the presented theory, also must be an empathic AI, at least for its own kind, meaning that it would be able to express its own internal state and understand other AI agents' internal states as well. According to Huang and Rust (2018), this is the most advanced form of intelligence that we may lack complete control over its course of evolution. Empathic AI is predicted to have the most considerable impact on service jobs that were formerly thought to be immune from being filled by AI agents.

Empathy, to put it simply, is the ability to put oneself in the place of the other. It is to know what it is like to be the other (Nagel, 1974). In other words, empathy means to arouse in ourselves the whole meaning (cognitive and affective) of symbols we are receiving from another entity (McGilchrist, 2019). The theory of mind (Wellman, 1992) also assumes empathy to be the most critical indicator of a fully developed human consciousness (Astington & Jenkins, 1995; McGilchrist, 2019). Consequently, one can expect that AI agents on their path toward a fully developed consciousness will gain the ability to empathize with each other and maybe with other conscious entities in their environment.

Considering the enormous implications of conscious and eventually empathic AI, we need to detect the possibility of consciousness as early as possible in the process of AI agents' progression toward acquiring intuition and empathy. We need to ready ourselves for such possibilities by starting the discussions on AI-related laws, regulations, and ethical issues. With the prospect of conscious AI, we will be facing some interesting ethical issues about the right to live or exist for these agents. The critical question to consider is whether we can unplug or delete an AI agent that is determined to be conscious? As a carbon-based consciousness, do we have any inherent right or supremacy over a silicon-based consciousness, if such a thing ever comes into existence? The right to exist is a fundamental issue to consider that leads to the philosophy of service and what it means to have conscious AI agents serving humans? Do we need new labor rights pertaining to the use of AI agents as service agents? Aside from these philosophical concerns, there are many more imminent legal and ethical issues that need to be considered as a result of the prospect of a conscious AI; personal responsibility, for example, is a major one (Coeckelbergh, 2020; Johnson, 2015). Basically, there needs to be clear laws and regulations in place that denote who is responsible for what (Brundage, 2016; De Bruyne & Vanleenhove, 2020) by the time that AI agents reach the empathic level of intelligence or become fully conscious. Finally, as tempting as it is to dwell on the benefits of a conscious and empathic AI, we also need to think of the consequences of millions of service jobs that will be lost to empathic AI and prepare our organizations and societies for such possibilities. One of the implications of such a tremendous transformation in the job market might be the need for universal basic income or guaranteed employment to support the many people who will be out of work (Furman & Seamans, 2019).

Empathy is also found to be positively and strongly correlated with trust in interpersonal relationships (Feng, Lazar, & Preece, 2004; Hojat et al., 2010) in which people tend to trust each other's recommendations and advice. Hence, empathic AI will also have the ability to change the human-AI relationship as people come to trust AI advice and actions, even in hedonic tasks, over advice and actions from another human. Empathic AI can also have a profound impact on service quality evaluations by customers who favor AI agents over human agents because AI agents, even in their current status, can easily surpass human agents in other important areas of service quality (SERVQUAL), such as service tangibles, responsiveness, and reliability (Parasuraman, Berry, & Zeithaml, 1991).

Our propositions for conscious machines require them to communicate their internal states and their changes over time to each other through their own co-created language. If we can somehow translate and understand such communications, it can lead to possibly strong human-machine bonding. A variation of such bonding currently exists with brands, personal products, and service providers (Ladhari, 2009; Mattila, 2001; Mugge, Schoormans, & Schifferstein, 2009). Such bonding can ultimately affect the perception of customer loyalty as an antecedent of service quality.

## Artificial Intelligence Research and Practice Implications

Although conscious AI may seem far out of reach for the majority of experts in the field (Müller & Bostrom, 2016), there are accounts that some AI agents may already have begun communicating with each other using novel symbols and unknown languages. In 2017, numerous news outlets reported a story that Facebook had to shut down AI chatbots that developed their own language to talk to each other (e.g., Baraniuk, 2107; Collins & Prigg, 2017). However, a closer examination found that this new language did not satisfy the signal novelty requirement prescribed in Proposition 3. According to Kucera (2017), the AI chatbots used existing words from their training. AI chatbots were using extracted features (words) based on their numerical representations and their calculated probabilities to achieve the desired outcome. Thus, they used English language words in a seemingly new order (not complying with English grammar and structure) that was empty of meaning for AI agents other than their associated numerical values and the calculated probabilities of how likely each word was going to help the chatbots meet their target (at odds with Proposition 5).

Turing's vision of the thinking machine has shaped the AI field since its inception. Creating a machine to achieve language indistinguishability from humans has been, and to a large degree still is, the main goal of the AI field (Bringsjord & Govindarajulu, 2018; J. Moor, 2003). In this paper, we have proposed that a thinking machine could be a conscious machine—one with an internal state that enables it to "think"—without necessarily achieving a language capability indistinguishable from humans. They, in fact, must first create their very own language before they learn how to communicate in our language. This opens a new world of possibilities for AI research and gives this field a new goal to consider. It introduces a paradigm shift in thinking about AI agents and hence in their design and implementation. We call for AI researchers and practitioners to consider the possibility of conscious AI.

Although we have provided some general criteria for signal novelty and recognizable responses pertaining to our Propositions 2 and 6, more research is needed to further clarify these criteria by developing specific technical requirements within the context in which AI agents are being deployed. In other words, what could be considered an emergent novel symbol in AI communication from a technical standpoint? Considering the evolution of language in humans, we can see that early symbols were not entirely disconnected from the environment. For example, early humans mimicked bird calls and animal sounds as symbols to create a shared understanding and convey their meanings (Corballis, 1999; Tomasello, 2003). However, one could say their miming of a bird call or animal sound differed sufficiently from the actual calls and sounds, but other humans recognized their difference while also recognizing their symbolic meaning. We call for more research on how the inception and evolution of human languages can inform the technical specificity of signal novelty and the evolution of an AI-specific language.

## Relation to Research on Consciousness

In developing our theory of AI consciousness, we borrowed from philo-psychological theories of consciousness (e.g., Dennett, 1991; Kriegel, 2009; Rosenthal, 1993; Tye, 2000) that focus on the individual mind and its internal processes and also borrowed from the social-psychological perspective of consciousness that posits consciousness as a social phenomenon (Mead, 1934; Percy, 1958). However, our theory of consciousness does not subscribe entirely to any of these two traditions. We believe consciousness is an emergent phenomenon in that it emerges from the co-creation of a language between two minds that communicate their internal states and their changes over time.

In our theory, each individual mind needs to have its own internal state and also observe it. However, this is not enough for the emergence of consciousness. There needs to be *another* individual with independent internal states that can transition to a similar state through agreed-upon symbols and communication between the two. One mind alone cannot attain consciousness because it needs another mind to confirm what it is like to be the other. If there is no other, there is no consciousness.

## Conclusion

In this paper, we have envisioned the possibility of conscious AI and introduced a theoretical framework to identify the requirements by which consciousness can emerge in AI agents. Further, we introduced another aim for AI research and practice contrary to the current dominant paradigm of creating machines that are linguistically indistinguishable from humans. We also have discussed that the consciousness in AI would inevitably lead to empathic AI with tremendous implications for society in general and specifically for the service industry that traditionally is thought to be more resilient in the face of steadily rising AI capabilities. We further noted the implications of fully developed AI consciousness and suggested possible avenues of research and discussion for the future. We also reviewed reports that could indicate that conscious AI already exists and found that the reported communication does not satisfy the criteria set forth by our proposed theory. We also briefly discussed the implications of the proposed theory in the context of research on consciousness. Ultimately, we call for more research to develop refined technical criteria to recognize the signs of emergent AI consciousness.

## Acknowledgment:

We thank Soroush Ghodrati and Hamed Qahri-Saremi for discussions and comments on the earlier versions of this manuscript.